\documentclass[10pt,twocolumn,letterpaper]{article}

\usepackage{cvpr}              

\usepackage{graphicx}
\usepackage{amsmath}
\usepackage{amssymb}
\usepackage{booktabs}
\usepackage{blindtext}

\definecolor{pretty-blue}{RGB}{0, 113, 188}
\definecolor{linecolor}{gray}{.89} 
\definecolor{linecolor3}{gray}{.95} 
\definecolor{linecolor2}{gray}{.94} 
\definecolor{linecolor1}{gray}{.96} 

\definecolor{pretraincolor}{HTML}{2E59A7} 

\definecolor{line3}{HTML}{F0FEF0}
\definecolor{line2}{HTML}{DCFCF9}
\definecolor{line1}{HTML}{B3F7F4}

\definecolor{mambacolor}{HTML}{6424D6}
\definecolor{reconcolor}{HTML}{412F8A}
\definecolor{resnetcolor}{HTML}{8DA0CB}
\definecolor{vitcolor}{HTML}{fc8e62}
\newcommand{\reconcolor}[1]{\textcolor{reconcolor}{#1}}
\newcommand{\vitcolor}[1]{\textcolor{vitcolor}{#1}}

\newcommand{\br}{\reconcolor{$\star$\,}} 
\newcommand{\bs}{\vitcolor{$\mathbf{\circ}$\,}} 
\newcommand{\bh}{\reconcolor{$\mathbf{\diamond}$\,}} 

\newcommand{\ours}{{\texttt{MoST}}\xspace}

\definecolor{newgreen}{HTML}{019C74}
\definecolor{berryred}{HTML}{ED656B}
\definecolor{chinablue}{HTML}{66A9C9}
\newcommand{\dminus}[1]{\fontsize{9pt}{0.1em}\selectfont (\textbf{\textcolor{berryred}{#1}})}
\newcommand{\dplus}[1]{\fontsize{9pt}{0.1em}\selectfont (\textbf{\textcolor{newgreen}{#1}})}
\newcommand{\dperc}[1]{\fontsize{9pt}{0.1em}\selectfont (\textbf{\textcolor{chinablue}{#1\%}})}

\usepackage{times}
\usepackage{epsfig}

\usepackage{multirow}
\usepackage{bbm}
\usepackage{multirow}
\usepackage{algorithm}
\usepackage[noend]{algpseudocode}
\usepackage{tabularx}
\usepackage{makecell}
\usepackage{microtype}
\usepackage{textcomp}
\usepackage{colortbl}
\usepackage{url}
\usepackage{cuted}

\usepackage{caption}
\usepackage[outline]{contour}

\usepackage{pifont}  

\definecolor{zimabule}{HTML}{272CE5}
\definecolor{bulered}{HTML}{6527fe}
\definecolor{greenblue}{HTML}{00acc1}
\definecolor{orangered}{HTML}{ed635e}
\definecolor{orange}{HTML}{ff4e00} 
\definecolor{lightred}{HTML}{ff3886} 
\definecolor{lightpurple}{HTML}{1000ca} 

\usepackage{wrapfig} 
\usepackage{ulem}
\usepackage{soul}

\definecolor{cvprblue}{rgb}{0.21,0.49,0.74}
\usepackage[pagebackref,breaklinks,colorlinks,allcolors=cvprblue]{hyperref}

\def\paperID{1869} 
\def\confName{CVPR}
\def\confYear{2025}

\title{\ours: Efficient Monarch Sparse Tuning for 3D Representation Learning}

\author{Xu Han \qquad Yuan Tang \qquad Jinfeng Xu \qquad Xianzhi Li$^\dag$\\
Huazhong University of Science and Technology\\
{\tt\small \{xhanxu, yuan\_tang, jinfengx, xzli\}@hust.edu.cn \quad $^\dag$Corresponding author.} 
}

\begin{document}
\maketitle

\begin{strip}
\begin{minipage}{\textwidth}\centering
\vspace{-3.4em}
\includegraphics[width=1.0\textwidth]{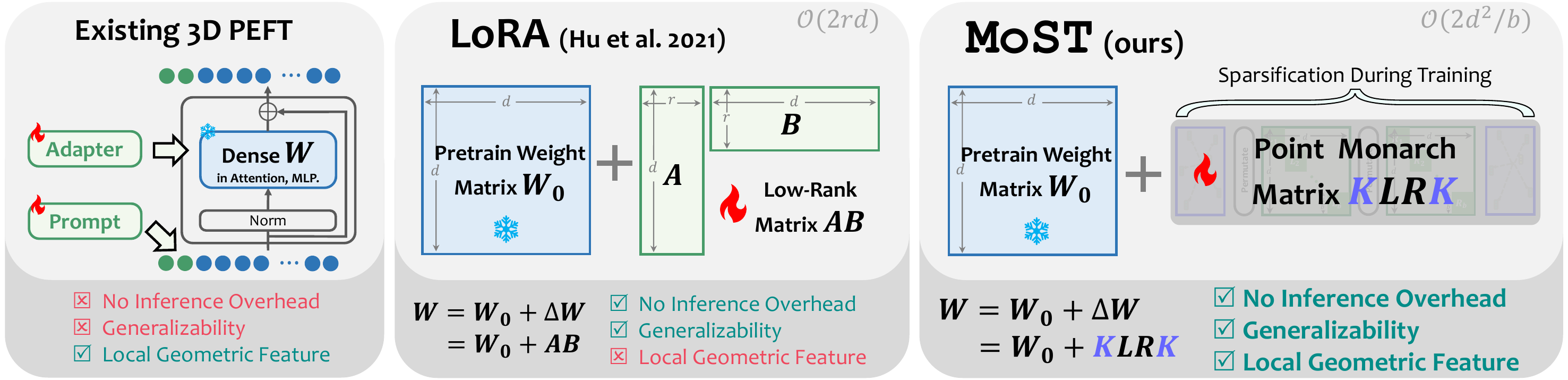}
\captionof{figure}{
Existing 3D parameter-efficient fine-tuning (PEFT) methods rely on additional adapters or prompts, which, while using point cloud priors, introduce inference overhead and lack generalization.
Reparameterization-based PEFT methods like LoRA\cite{LoRA}, though free of the above issues, overlook point cloud characteristics.
\ours combines the best of both worlds by reparameterizing dense update weight matrices with tailored sparse Point Monarch matrices, preserving local geometry, avoiding inference overhead, and remaining generalizable.
}
\label{fig:teaser}
\end{minipage}
\end{strip}

\begin{abstract}

%
We introduce Monarch Sparse Tuning (\ours), the first reparameterization-based parameter-efficient fine-tuning (PEFT) method tailored for 3D representation learning.
Unlike existing adapter-based and prompt-tuning 3D PEFT methods, \ours introduces no additional inference overhead and is compatible with many 3D representation learning backbones.
At its core, we present a new family of structured matrices for 3D point clouds, Point Monarch, which can capture local geometric features of irregular points while offering high expressiveness. 
\ours reparameterizes the dense update weight matrices as our sparse Point Monarch matrices, significantly reducing parameters while retaining strong performance.
Experiments on various backbones show that \ours is simple, effective, and highly generalizable. 
It captures local features in point clouds, achieving state-of-the-art results on multiple benchmarks, e.g., 97.5\% acc. on ScanObjectNN (PB\_50\_RS) and 96.2\% on ModelNet40 classification, while it can also combine with other matrix decompositions (e.g., Low-rank, Kronecker) to further reduce parameters.

\end{abstract}    
\section{Introduction}
\label{sec:intro}

%
3D point cloud analysis finds extensive applications in autonomous driving~\cite{qi2021offboard, PointRCNN}, embodied AI~\cite{Point-Cloud-Matters}, and virtual reality~\cite{DL4Point}.
In recent years, learning-based point cloud analysis has advanced rapidly, achieving leading performance across various tasks.
Pioneered by PointNet~\cite{PointNet}, most point cloud learning models focus on designing intricate models to learn geometric or semantic features from unstructured points~\cite{PointNet++, DGCNN, PointCNN, KPConv, PointNext}.
Inspired by the success of large-scale pretraining in language and image modeling~\cite{GPT3_20, BERT, MAE}, recent approaches in point cloud learning~\cite{PointBERT, PointMAE, PointGPT, PointMamba, Mamba3D, PointM2AE22} leverage pretrained backbones like Transformer~\cite{AttentionIsAllYouNeed}, Mamba~\cite{Mamba}, and hierarchical architectures~\cite{U-Net} for 3D representation learning. 
This pretrain-finetune paradigm leads to more robust point cloud features, boosting performance on downstream tasks~\cite{Point_Cloud_Learning_Survey}.

%
A problem arises when applying pretrained models to downstream tasks: the entire model needs to be retrained on downstream datasets, requiring substantial computation and memory---impractical for many tasks.
To address this, recent efforts on Parameter-Efficient Fine-Tuning (PEFT)~\cite{PEFT-survey} aim to achieve strong performance with minimal parameter updates.
PEFT methods fall into three categories: additive (e.g., Adapters~\cite{Adapter}, Prompt Tuning~\cite{Prompt-Tuning}), selective (e.g., BitFit~\cite{BitFit}), and reparameterization-based (e.g., LoRA~\cite{LoRA}). 
Additive methods introduce new model components or input tokens, training only these additional parameters. 
Selective methods fine-tune a specific subset of the model. 
Reparameterization-based methods replace dense update weight matrices with sparse approximations, like the low-rank approximation used in LoRA.
Given that most models contain numerous dense layers, such replacements offer a superior balance between generalization and efficiency.

\begin{figure}[t]
    \begin{center}
    \includegraphics[width=1.0\linewidth]{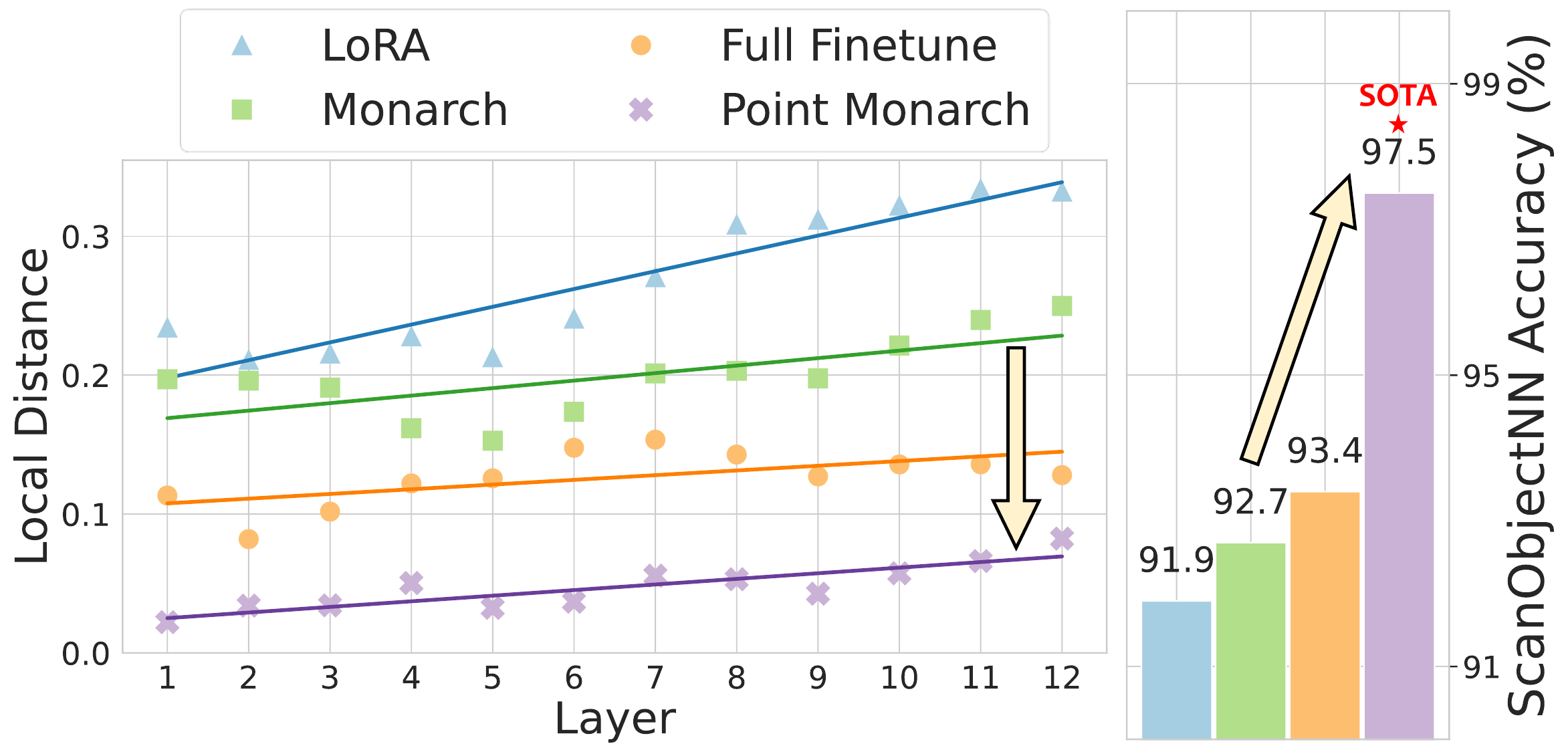}
    \vspace{-17pt}
    \caption{
    The average $l_2$-distance of features between KNN centers and neighbors after applying different structured matrices (left). We observe this local feature distance correlates with classification acc. on PB\_50\_RS~\cite{ScanObjectNN19} (right).
    LoRA and Monarch~\cite{Monarch} lead to higher distances, Point Monarch exhibits the lowest one by smoothing local geometric features, achieving 97.5\% acc. with PointGPT~\cite{PointGPT}.
    }\label{fig:local_dist}
    \end{center}
    \vspace{-18pt}
\end{figure}

%
Existing PEFT methods primarily target language and image modeling~\cite{Delta-Tuning, PEFT-survey}, often falling short when applied to 3D point clouds.
The field of 3D PEFT remains underexplored, with most efforts focusing on additive methods that use custom adapters or prompts tailored for Transformer~\cite{IDPT, DAPT, PPT, Point-PEFT}.
While these methods significantly reduce the fine-tuning parameters and effectively incorporate point cloud priors, such as local features, they face two key limitations: 
(1) Limited generalization and flexibility: adapters or prompts designed for Transformer~\cite{AttentionIsAllYouNeed} do not generalize well to other backbones like Mamba~\cite{Mamba} or hierarchical models~\cite{U-Net}, and adjusting parameter sizes is difficult, hindering scalability. 
(2) They introduce additional inference overhead and often fail to match the performance of full fine-tuning.
Intuitively, a reparameterization-based PEFT method offers a solution by preserving the model architecture and avoiding inference overhead, as it simply reparameterizes dense weight matrices with sparse ones during training.
However, in practice, methods like LoRA~\cite{LoRA}, which rely on low intrinsic rank, tend to capture global information but struggle with local feature learning. As shown in \cref{fig:local_dist}, this limitation severely impacts 3D PEFT performance.

%
To fill this gap, we propose {Monarch Sparse Tuning} (\ours), the first reparameterization-based 3D PEFT method.
At its core, we introduce Point Monarch, a novel structured matrix family tailored for 3D point clouds, extending Monarch~\cite{Monarch} to unstructured points.
Point Monarch captures local geometric features with linear transformations, and features a block-wise structure that aligns well with patch-based point cloud learning, outperforming low-rank matrices.
\ours replaces dense update weight matrices with Point Monarch during training, supporting various backbones without adding inference overhead.
\ours captures local features effectively and is highly expressive, enabling it to match or exceed full fine-tuning across multiple tasks.
Additionally, \ours allows users to balance efficiency and performance based on task needs, and supports further parameter reduction via matrix decompositions, such as Low-rank and Kronecker~\cite{Kronecker}.
Finally, we design a parameter-free multi-layer feature fusion strategy, to boost knowledge transfer from the backbone to the task header, avoiding bottlenecks.

%
We validate \ours across diverse settings, including masked point, multimodal, and large-scale pretraining, using various backbones such as Transformers, Mamba, and hierarchical architectures.
Extensive experiments show that \ours consistently outperforms existing PEFT methods in object- and scene-level classification and segmentation tasks, even surpassing full fine-tuning in almost all classification tasks while tuning just 3.6\% of parameters.
With \ours, Point-MAE~\cite{PointMAE} and I2P-MAE~\cite{I2P-MAE} achieve accuracy gains of 7.74\% and 3.16\%, respectively, over full fine-tuning on the ScanObjectNN~\cite{ScanObjectNN19} (PB\_50\_RS). 
Mamba3D~\cite{Mamba3D} and PointGPT~\cite{PointGPT} achieve 95.2\% and \textbf{96.2\%} accuracy on ModelNet40~\cite{ModelNet15}, and 93.3\% and \textbf{97.5\%} on PB\_50\_RS. 
\ours fine-tuned ReCon~\cite{ReCon} surpasses the state-of-the-art 3D PEFT method by 0.2\% class mIoU on ShapeNetPart~\cite{ShapeNetPart16} part segmentation and 1.0\% mIoU on S3DIS~\cite{S3DIS16} scene segmentation.

%
In summary, our main contributions are as follows.
\begin{itemize}
\item We introduce {Point Monarch}, a novel structured matrix family tailored for 3D point clouds that captures both global and local features while maintaining sparsity.
\item We propose {\ours}, the first reparameterization-based 3D PEFT method, which generalizes well across different backbones without adding inference overhead. 
\item Extensive results across various models demonstrate \ours's generalization and effectiveness, achieving {state-of-the-art} 3D PEFT on object- and scene-level tasks. 
\end{itemize}

\begin{figure*}[t]
    \begin{center}
    \vspace{-4pt}
    \includegraphics[width=1.0\linewidth]{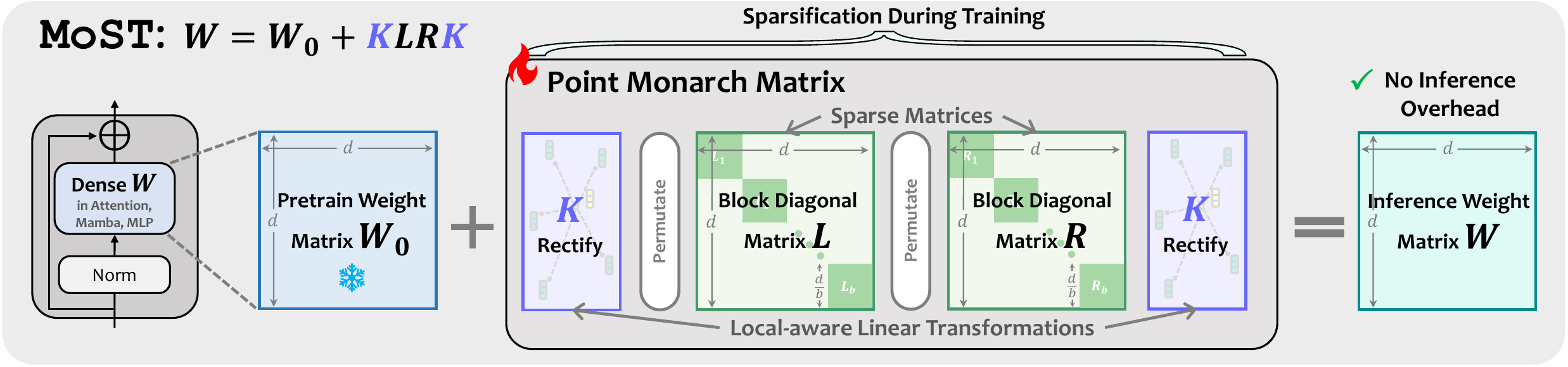}
    \vspace{-15pt}
    \caption{\textbf{Illustration of Monarch Sparse Tuning}. During training, \ours reparameterizes dense update weight matrices using our sparse and expressive Point Monarch matrices \( \mathbf{\textcolor{mambacolor}{K} L R \textcolor{mambacolor}{K}} \), which capture local geometric features of points through simple linear transformations.
    }\label{fig:framework}
    \vspace{-20pt}
    \end{center}
\end{figure*}

\section{Related Work}
\label{sec:related_work}

\subsection{3D Representation Learning}
3D representation learning~\cite{Point_Cloud_Learning_Survey} aims to extract robust point cloud features through pretraining on large datasets, boosting performance in downstream tasks.
Inspired by Masked Language Modeling~\cite{BERT} and Masked Image Modeling~\cite{MAE}, many point cloud pretraining methods mask parts of the point cloud and train models to reconstruct these masked regions~\cite{Point_Cloud_Learning_Survey}.
Notable methods like Point-BERT~\cite{PointBERT} and Point-MAE~\cite{PointMAE} leverage this to learn effective representations, excelling in tasks such as classification and segmentation.
Recently, with the increasing availability of 3D-text data pairs, multimodal pretraining methods like ACT~\cite{ACT23} and ReCon~\cite{ReCon, ReCon++} have shown superior effectiveness.
However, fine-tuning  the entire pretrained models for downstream tasks is resource-intensive, calling for effective 3D Parameter-Efficient Fine-Tuning (PEFT)~\cite{PEFT-survey}.
Here, we introduce the first reparameterization-based 3D PEFT method, specifically designed for irregular 3D point clouds.

\subsection{3D Parameter-Efficient Fine-Tuning}
Current 3D PEFT primarily focuses on adapter~\cite{Adapter} and prompt tuning~\cite{Prompt-Tuning}.
Pioneers like IDPT~\cite{IDPT} and DAPT~\cite{DAPT} explore this approach: IDPT employs DGCNN~\cite{DGCNN} as an adapter to generate prompts for Transformers, while DAPT uses a dynamic adapter that combines adapter and prompt tuning.
Both methods significantly reduce trainable parameters, leveraging point cloud features to enhance performance. 
Recently, PPT~\cite{PPT} demonstrates the effectiveness of fine-tuning position encodings, and PointGST~\cite{PointGST} proposes an adapter to extract spectral-domain features for efficient 3D PEFT.
However, these adapter-based and prompt tuning methods~\cite{Point-PEFT, sun2024parameter, Any2point, fei2024fine, fei2024parameter} introduce inference overhead and are specifically designed for Transformers~\cite{AttentionIsAllYouNeed}, limiting adaptability to other architectures like Mamba~\cite{Mamba} and U-Net~\cite{U-Net}.
In contrast, our \ours overcomes these limitations from the perspective of reparameterization, avoiding inference overhead while maintaining high generalizability.

\subsection{Structured Matrices}
Dense matrices exhibit quadratic complexity, whereas structured matrices offer sub-quadratic parameter counts and runtime~\cite{Compute_Better_Spent}, making them ideal for replacing dense layers in large models to reduce computation.
Common structured matrices include Low-rank, Kronecker, and Toeplitz matrices~\cite{Toeplitz} (or known as convolutions), as well as various fast transforms such as Fourier and sine/cosine.
Notably, inspired by the divide-and-conquer scheme from \citet{de2018two}, \citet{Butterfly} introduce Butterfly matrices, which are highly expressive but inefficient on modern GPUs.
Monarch matrices~\cite{Monarch} then extend Butterfly by leveraging optimized batch-matrix-multiply to accelerate factorization.
However, these structured matrices often struggle to capture local point cloud features, limiting their effectiveness in 3D PEFT.
In contrast, our Point Monarch captures local geometric features using \textit{two simple KNN-wise local token linear transformations, seamlessly integrating with efficient batch matrix multiplication} while maintaining a block-diagonal structure suited for patch-based point cloud learning.

\section{Method}

\begin{figure*}[t]
    \begin{center}
    \vspace{-4pt}
    \includegraphics[width=1.0\linewidth]{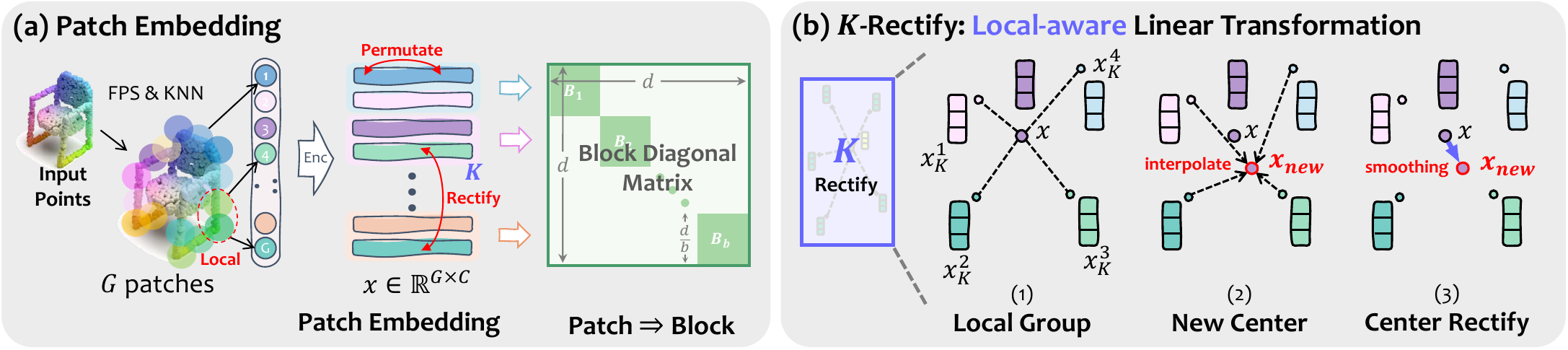}
    \vspace{-15pt}
    \caption{\textbf{Illustration of Patch Embedding and \( K \)-Rectify}. (a) Our Point Monarch performs channel-wise permutation and token-wise local rectification, and its block-wise structure aligns with patch-based point cloud representation learning. (b) \( K \)-Rectify groups local features based on xyz coordinates and interpolates a new center feature, then facilitating the center feature rectification.
    }\label{fig:k_rectify}
    \vspace{-20pt}
    \end{center}
\end{figure*}

During training, \ours (\cref{sec:most}) reparameterizes dense update weight matrices into sparse Point Monarch (\cref{sec:point_monarch}), extending Monarch (\cref{sec:monarch}) to point cloud learning, capturing local features while maintaining high expressiveness.
We also propose a feature fusion strategy (\cref{sec:fusion}) to enhance the backbone-header alignment without learning.

\subsection{Preliminaries: Monarch Matrices}
\label{sec:monarch}
\citet{Monarch} introduce {Monarch} to enhance hardware efficiency over Butterfly~\cite{Butterfly}. Following sparse matrix product factorization~\cite{de2018two}, a Monarch matrix $\mathbf{M}$ is defined as:
\begin{equation}
\mathbf{M} = \mathbf{P L P^\top R},
\end{equation}
where $\mathbf{P}$ is a permutation from row-major to column-major, commonly used in CNNs like ShuffleNet~\cite{Shufflenet}. $\mathbf{L}$ and $\mathbf{R}$ are sparse block-diagonal matrices of size $d$ with $b$ blocks, where each block is of size $d / b$. Typically, with $b \approx \sqrt{d}$, Monarch’s total parameter count is $2d^2 / b \ll d^2$.

As shown in \cref{fig:local_dist}, we observe that the average $l_2$-distance of features between KNN centers and neighbors correlates with classification accuracy on ScanObjectNN~\cite{ScanObjectNN19}, with lower distances indicating higher accuracy. We assume that this local feature distance reflects the model's ability to capture local features. Both LoRA and Monarch show higher distances than full fine-tuning, leading to lower accuracy. Despite Monarch's higher expressiveness compared to LoRA~\cite{LoRA}, it still struggles to capture local geometric features in point clouds, limiting its performance in 3D PEFT tasks. Inspired by this, we design Point Monarch to reduce the local feature distance and optimize performance.

\subsection{Point Monarch}
\label{sec:point_monarch}
Following the divide-and-conquer factorization scheme~\cite{de2018two}, we introduce {Point Monarch}, a structured matrix family tailored for 3D point clouds, as shown in \cref{fig:framework}, defined as:
\begin{equation}
    \text{Point Monarch} = \mathbf{\textcolor{mambacolor}{K} P L P^\top R \textcolor{mambacolor}{K}},
\end{equation}
where \( \textcolor{mambacolor}{\mathbf{K}} \), representing \( K \)-Rectify, facilitates information exchange among tokens within geometric (xyz) neighborhoods, differing from channel information fusion with permutation $\mathbf{P}$. 
$\mathbf{L}$ and $\mathbf{R}$ are block-diagonal matrices of size $d \times d$, with $b$ blocks of $\mathbf{L_i / R_i} \in \mathbb{R}^{\frac{d}{b} \times \frac{d}{b}}$ on the diagonal: 
\begin{align}
\mathbf{L} &= \text{diag($\mathbf{L_1}$, $\mathbf{L_2}$,..., $\mathbf{L_b}$)},  \\
\mathbf{R} &= \text{diag($\mathbf{R_1}$, $\mathbf{R_2}$,..., $\mathbf{R_b}$)}.
\end{align}

Point Monarch supports efficient batch matrix multiplication, making it hardware-friendly. 
It balances efficiency, expressiveness, and local feature capturing, with a parameter count of $2d^2 / b \ll d^2$ where typically $b \approx \sqrt{d}$ to control sparsity. 
Replacing dense layers with sparse Point Monarch improves parameter efficiency, similar to model pruning~\cite{Deep_Compression}, and actually can be seen as a handcrafted lottery ticket structure~\cite{Lottery_Ticket} tailored for point cloud processing. 
Here, \( \textcolor{mambacolor}{\mathbf{K}} \) serves as \textit{a KNN-wise local token linear transformation}, \textit{integrating seamlessly with batch matrix multiplication while preserving the block-diagonal Monarch structure}.
For simplicity, we abbreviate Point Monarch as \( \mathbf{\textcolor{mambacolor}{K} L R \textcolor{mambacolor}{K}} \).

\vspace{5pt}
\noindent\textbf{\( K \)-Rectify.} 
As shown in \cref{fig:k_rectify}, \( \textcolor{mambacolor}{\mathbf{K}} \) rectifies patch embeddings within local neighborhoods (\cref{fig:k_rectify}a) through three steps (\cref{fig:k_rectify}b): KNN local grouping, new center token interpolation, and rectification. 
For a token sequence \( \mathbf{x} \in \mathbb{R}^{G \times C} \), where \( G \) is the sequence length, and \( C \) is the number of feature channels, \( \textcolor{mambacolor}{\mathbf{K}} \) (\( K \)-Rectify) is defined as:
\begin{align}
\mathbf{x}_K &= \text{KNN}(\mathbf{p}, \mathbf{x}), \\
\mathbf{x}_{\text{new}} &= \text{IDW}(\mathbf{x}_K), \\
\textcolor{mambacolor}{\mathbf{K}} \mathbf{x} &= \mathbf{x} + \lambda \mathbf{x}_{\text{new}}.
\end{align}
Here, \( \mathbf{x} \) contains embeddings for center points \( \mathbf{p} \in \mathbb{R}^{G \times 3} \) across \( G \) patches. 
The local embedding groups \( \mathbf{x}_K \in \mathbb{R}^{G \times K \times C} \) are obtained by KNN to find geometric neighbors of \( \mathbf{p} \), and grouping the corresponding embeddings, with group size \( K \). 
We compute new center embeddings \( \mathbf{x}_{\text{new}} \in \mathbb{R}^{G \times C} \) using inverse distance weighting (IDW) of local groups $\mathbf{x}_K$. 
Finally, \( \mathbf{x}_{\text{new}} \) refines the old \( \mathbf{x} \) with a hyperparameter $\lambda$, resulting in the rectified token sequence $\textcolor{mambacolor}{\mathbf{K}} \mathbf{x}$. 
The IDW is formulated as:
\begin{equation}
    \text{IDW}(\mathbf{x}_K) = \sum_{i=1}^K \frac{\frac{1}{\| p - p_i \|^2}}{\sum_{j=1}^K \frac{1}{\| p - p_j \|^2}} \cdot x_K^i,  \quad i = 1, \dots, K
\end{equation}
where $p$ is one of the center points $\mathbf{p}$, and $p_i$ is one of $p$'s KNN geometric neighbors.

\vspace{5pt}
\noindent\textbf{Matrix Form.} 
Given the adjacency matrix $\mathbf{A} \in \mathbb{R}^{G \times G}$, where $\mathbf{A}_{ij} = 1$ if points $i$ and $j$ are KNN neighbors, otherwise $0$, and the distance matrix $\mathbf{D} \in \mathbb{R}^{G \times G}$, where $\mathbf{D}_{ij}$ is the normalized inverse distance $1/d_{ij}$ between points $i$ and $j$, we express \( \textcolor{mambacolor}{\mathbf{K}} \in \mathbb{R}^{G \times G} \) as a \textit{linear transformation matrix}:
\[
\textcolor{mambacolor}{\mathbf{K}} = \mathbf{I} + \lambda {\mathbf{A}} \odot \mathbf{D},
\]
where $\odot$ is the Hadamard product, $\mathbf{I} \in \mathbb{R}^{G \times G} $ is the identity matrix. See Appendix for proof. Note that $\textcolor{mambacolor}{\mathbf{K}}$ is also sparse.

\vspace{5pt}
\noindent\textbf{Key Insight.} 
Local features in point clouds are best captured by fusing information within KNN neighborhoods \cite{PointNet, PointNet++}. 
This fusion smooths local features while preserving their distinctiveness. 
In point cloud representation learning, it facilitates interactions among tokens in KNN neighborhoods, balancing local and global feature learning. 
While MLPs, Attention~\cite{AttentionIsAllYouNeed}, or Mamba~\cite{Mamba} can capture global features, our \( K \)-Rectify enhances local feature fusion through two simple KNN-wise linear transformations, smoothing local features while remaining hardware efficiency.

\vspace{5pt}
\noindent\textbf{Expressiveness.} 
\citet{de2018two} and \citet{Butterfly, Monarch} demonstrate that all structured matrices can be expressed as products of Butterfly or Monarch, achieving optimal memory and runtime complexity with polylogarithmic factors. 
Since Point Monarch extends Monarch to points, it inherits its high expressiveness, near-optimal runtime, and parameter efficiency. 
Furthermore, Point Monarch smooths local geometric features, making it more effective for point clouds.

\begin{table*}[t!]
\renewcommand{\arraystretch}{1.05}
\caption{
\textbf{PEFT Methods on Various Models}. We report the number of trainable parameters \#P (M), ScanObjectNN PB\_50\_RS~\cite{ScanObjectNN19} (SO) accuracy (\%) w/o voting, and ModelNet40~\cite{ModelNet15} (MN) accuracy (\%) w/o voting. We compare different PEFT methods across \bh hierarchical architectures, \bs Transformers, and \br Mamba-based models for 3D understanding. \texttt{3D}: designed for 3D PEFT. \texttt{NO}: no inference overhead.
}
\label{tab:peft_main}
\begin{center}
\vspace{-15pt}
\resizebox{\linewidth}{!}{
\begin{tabular}{lcccccccccccc}
\toprule[0.7pt]
\multirow{2}{*}[-0.5ex]{\shortstack{Method}} & \multirow{2}{*}[-0.5ex]{\shortstack{\texttt{3D}}} & \multirow{2}{*}[-0.5ex]{\shortstack{\texttt{NO}}} & \multicolumn{2}{c}{\bs Point-MAE~\cite{PointMAE}} & \multicolumn{2}{c}{\bh I2P-MAE~\cite{I2P-MAE}} & \multicolumn{2}{c}{\bs ReCon~\cite{ReCon}} & \multicolumn{2}{c}{\br Mamba3D~\cite{Mamba3D}} & \multicolumn{2}{c}{\bs PointGPT~\cite{PointGPT}} \\
\cmidrule(lr){4-5} \cmidrule(lr){6-7} \cmidrule(lr){8-9} \cmidrule(lr){10-11} \cmidrule(lr){12-13}
& & & \#P $\downarrow$ & SO\ /\ MN $\uparrow$  & \#P $\downarrow$ & SO\ /\ MN $\uparrow$  & \#P $\downarrow$ & SO\ /\ MN $\uparrow$  & \#P $\downarrow$ & SO\ /\ MN $\uparrow$ & \#P $\downarrow$ & SO\ /\ MN $\uparrow$ \\
\midrule[0.4pt]
Full FT & - & - & 22.1 & 85.18\ /\ 93.8 & 15.3 & 90.11\ /\ 93.7 & 22.1 & 90.01\ /\ 92.5 & 16.9 & 92.05\ /\ 94.7 & 360.5 & 93.4\ /\ 94.1 \\
\midrule[0.4pt]
\multicolumn{13}{c}{\textit{Additive-based Parameter-Efficient Fine-Tuning} }\\
\midrule[0.4pt]
VPT~\cite{VPT22} & {\ding{55}} & {\ding{55}} & 0.4 & 81.09\ /\ 92.54 & 0.3 & 76.30\ /\ 89.10 & 0.4 & 84.04\ /\ 92.59 & 0.4 & 80.60\ /\ 91.13 & 1.1 & 91.60\ /\ 92.0 \\
IDPT~\cite{IDPT} & {\textbf{\textcolor{newgreen}{\ding{51}}}} & {\ding{55}} & 1.7 & 88.34\ /\ 93.64 & 1.6 & 85.70\ /\ 93.19 & 1.7 & 88.13\ /\ 93.64 & 1.7 & 87.34\ /\ 93.23 & 10.0 & 92.99\ /\ 93.4 \\
DAPT~\cite{DAPT} & {\textbf{\textcolor{newgreen}{\ding{51}}}} & {\ding{55}} & 1.1 & 88.27\ /\ 92.99 & 0.8 & {89.04}\ /\ {93.56} & 1.1 & 89.31\ /\ 93.27 & 1.1 & 88.55\ /\ 92.87 & 4.2 & 93.02\ /\ 94.2 \\
PPT~\cite{PPT} & {\textbf{\textcolor{newgreen}{\ding{51}}}} & {\ding{55}} & 1.1 & 89.00\ /\ {93.68} & 1.6 & 87.54\ /\ 92.99 & 1.1 & {89.52}\ /\ {93.76} & 1.1 & 87.61\ /\ 92.87 & 2.8 & 94.34\ /\ 94.2 \\
PointGST~\cite{PointGST} & {\textbf{\textcolor{newgreen}{\ding{51}}}} & {\ding{55}} & 0.6 & {89.3}\ /\ 93.5 & 0.5 & 87.27\ /\ 93.35 & 0.6 & 89.49\ /\ 93.6 & 0.6 & {89.97}\ /\ {93.72} & 2.4 & {94.83}\ /\ {94.8} \\
\midrule[0.4pt]
\multicolumn{13}{c}{\textit{Selective or Reparameterization-based Parameter-Efficient Fine-Tuning} }\\
\midrule[0.4pt]
BitFit~\cite{BitFit} & {\ding{55}} & {\textbf{\textcolor{newgreen}{\ding{51}}}} & 0.3 & 82.62\ /\ 92.42 & 0.3 & 85.18\ /\ 92.54 & 0.3 & 85.15\ /\ 92.34 & 0.3 & 88.97\ /\ 92.67 & 0.9 & 91.64\ /\ 92.79 \\
LoRA~\cite{LoRA} & {\ding{55}}  & {\textbf{\textcolor{newgreen}{\ding{51}}}} & 0.9 & 82.76\ /\ 92.50 & 0.8 & 84.00\ /\ 92.59 & 0.9 & 85.70\ /\ 92.87 & 0.9 & 87.16\ /\ 92.42 & 2.4 & 91.92\ /\ 92.95 \\
\rowcolor{linecolor}\textbf{\ours}$_{b=32}$ & {\textbf{\textcolor{newgreen}{\ding{51}}}} & {\textbf{\textcolor{newgreen}{\ding{51}}}} & 0.8 & \textbf{91.95}\ /\ \textbf{94.04} & 0.6 & \textbf{91.22}\ /\ \textbf{94.73} & 0.8 & \textbf{92.02}\ /\ \textbf{94.29} & 0.8 & \textbf{91.50}\ /\ \textbf{94.00} & 2.5 & \textbf{97.19}\ /\ \textbf{95.95} \\
\rowcolor{linecolor1}\textbf{\ours}$_{b=16}$ & {\textbf{\textcolor{newgreen}{\ding{51}}}} & {\textbf{\textcolor{newgreen}{\ding{51}}}} & 1.3 & \textbf{92.71}\ /\ \textbf{94.49} & 0.8 & \textbf{92.23}\ /\ \textbf{95.06} & 1.3 & \textbf{92.85}\ /\ \textbf{94.69} & 1.2 & \textbf{92.85}\ /\ \textbf{94.65} & 4.4 & \textbf{97.26}\ /\ \textbf{96.11} \\
\rowcolor{linecolor}\textbf{\ours}$_{b=8}$ & {\textbf{\textcolor{newgreen}{\ding{51}}}} & {\textbf{\textcolor{newgreen}{\ding{51}}}} & 2.3 & \textbf{92.92}\ /\ \textbf{94.77} & 1.4 & \textbf{93.27}\ /\ \textbf{95.14} & 2.3 & \textbf{93.55}\ /\ \textbf{95.06} & 2.1 & \textbf{93.30}\ /\ \textbf{95.18} & 8.0 & \textbf{97.50}\ /\ \textbf{96.23} \\
\bottomrule[0.7pt]
\end{tabular}
}
\end{center}
\vspace{-20pt}
\end{table*}

\begin{table*}[t!]
\renewcommand{\arraystretch}{1.06}
\caption{
\textbf{Results of Classification on ScanObjectNN PB\_50\_RS~\cite{ScanObjectNN19} and ModelNet40~\cite{ModelNet15}, Few-Shot Learning on ModelNet40, and Part Segmentation on ShapeNetPart~\cite{ShapeNetPart16}}. We report trainable parameters \#P (M), inference FLOPs \#F (G), overall accuracy (\%), class mIoU (\%), and instance mIoU (\%). We compare methods using \bh hierarchical architectures, \bs Transformers, and \br Mamba-based models for 3D understanding. $\dagger$: with voting strategy. Note that \ours uses Point Monarch with $b=16$ by default. All PEFT results are without voting.
}
\label{tab:cls_main}
\begin{center}
\vspace{-15pt}
\resizebox{\textwidth}{!}{
\begin{tabular}{lcccccccccc}
\toprule[0.7pt]
\multirow{2}{*}[-0.5ex]{Method} & \multicolumn{4}{c}{Shape Classification} & \multicolumn{4}{c}{Few-Shot Learning} & \multicolumn{2}{c}{Part Segmentation} \\
\cmidrule(lr){2-5} \cmidrule(lr){6-9} \cmidrule(lr){10-11}
& \#P $\downarrow$ & \#F $\downarrow$ & PB\_50\_RS $\uparrow$ & ModelNet40 $\uparrow$ & 5w10s $\uparrow$ & 5w20s $\uparrow$ & 10w10s $\uparrow$ & 10w20s $\uparrow$ & mIoU$_C$ $\uparrow$ & mIoU$_I$ $\uparrow$ \\
\midrule[0.4pt]
\multicolumn{11}{c}{\textit{Supervised Learning Only ({Dense Training})}}\\
\midrule[0.4pt]
\bh PointNet~\cite{PointNet} & 3.5 & 0.5 & 68.0 & 89.2 & 52.0 {\small $\pm$ 3.8} &  57.8 {\small $\pm$ 4.9} &  46.6 {\small $\pm$ 4.3} & 35.2 {\small $\pm$ 4.8} & 80.4 & 83.7 \\
\bh PointNet++~\cite{PointNet++} & 1.5 & 1.7 & 77.9 & 90.7 & 38.5 {\small $\pm$ 16.0} &  42.4 {\small $\pm$ 14.2} &  23.1 {\small $\pm$ 7.0} & 18.8 {\small $\pm$ 5.4} & 81.9 & 85.1 \\
\bh DGCNN~\cite{DGCNN} & 1.8 & 2.4 & 78.1 & 92.9 & 31.6 {\small $\pm$ 2.8} &  40.8 {\small $\pm$ 4.6} &  19.9 {\small $\pm$ 2.1} & 16.9 {\small $\pm$ 1.5} & 82.3 & 85.2 \\
\bs Transformer~\cite{AttentionIsAllYouNeed} & 22.1 & 4.8 & 77.2 & 91.4 & 87.8 {\small $\pm$ 5.2} & 93.3 {\small $\pm$ 4.3} & 84.6 {\small $\pm$ 5.5} & 89.4 {\small $\pm$ 6.3} & 83.4 & 85.1 \\
\br Mamba3D~\cite{Mamba3D} & 16.9 & 3.9 & {91.8} & {93.4} & {92.6 {\small $\pm$ 3.7}} & {96.9 {\small $\pm$ 2.4}} & {88.1 {\small $\pm$ 5.3}} & {93.1 {\small $\pm$ 3.6}} & {83.7} & {85.7} \\
\midrule[0.4pt]
\multicolumn{11}{c}{\textit{Supervised Learning Only ({Sparse Training})}}\\
\midrule[0.4pt]
\rowcolor{linecolor1}\bs \textbf{\ours} {\footnotesize (Transformer)} & 5.6 & 2.8 & \textbf{87.47} & \textbf{92.83} & \textbf{92.0 {\small $\pm$ 5.8}} & \textbf{95.4 {\small $\pm$ 2.9}} & \textbf{84.3 {\small $\pm$ 4.5}} & \textbf{88.8 {\small $\pm$ 4.3}} & \textbf{82.93} & \textbf{85.08} \\ 
\rowcolor{linecolor}\br \textbf{\ours} {\footnotesize (Mamba3D)} & 5.1 & 2.6 & \textbf{90.56} & \textbf{92.91} & \textbf{94.4 {\small $\pm$ 2.2}} & \textbf{96.0 {\small $\pm$ 2.6}} & \textbf{88.3 {\small $\pm$ 5.2}} & \textbf{89.4 {\small $\pm$ 4.7}} & \textbf{83.11} & \textbf{85.21} \\
\midrule[0.4pt]
\multicolumn{11}{c}{\textit{with Pre-Training (Full Fine-Tuning)}}\\
\midrule[0.4pt]
\bs Point-BERT~\cite{PointBERT} & 22.1 & 4.8 & 83.07 & 93.2 & 94.6 {\small $\pm$ 3.1} & 96.3 {\small $\pm$ 2.7} &  91.0 {\small $\pm$ 5.4} & 92.7 {\small $\pm$ 5.1} & 84.1 & 85.6 \\
\bs Point-MAE~\cite{PointMAE} & 22.1 & 4.8 & 85.18 & 93.8 & 96.3 {\small $\pm$ 2.5}&97.8 {\small $\pm$ 1.8} & {92.6 {\small $\pm$ 4.1}} & {95.0 {\small $\pm$ 3.0}} & 84.2 & 86.1 \\
\bh Point-M2AE~\cite{PointM2AE22} & 15.3 & 3.6 & 86.43 & 94.0 & 96.8 {\small $\pm$ 1.8} & 98.3 {\small $\pm$ 1.4} & 92.3 {\small $\pm$ 4.5} & 95.0 {\small $\pm$ 3.0} & 84.9 & 86.7 \\
\bh I2P-MAE~\cite{I2P-MAE} & 15.3 & 3.6 & 90.11 & 93.7 & 97.0 {\small $\pm$ 1.8} & 98.3 {\small $\pm$ 1.3} & 92.6 {\small $\pm$ 5.0} & 95.5 {\small $\pm$ 3.0} & {85.2} & {86.8} \\
\bs ACT~\cite{ReCon} & 22.1 & 4.8 & 88.21 & 93.7 & 96.8 {\small $\pm$ 2.3} & 98.0 {\small $\pm$ 1.4} & 93.3 {\small $\pm$ 4.0} & 95.6 {\small $\pm$ 2.8} & 84.7 & 86.1 \\
\bs ReCon~\cite{ReCon} & 43.6 & 5.3 &  90.63 & 94.1 & 97.3 {\small $\pm$ 1.9} & 98.9 {\small $\pm$ 1.2} & 93.3 {\small $\pm$ 3.9} & 95.8 {\small $\pm$ 3.0} & 84.8 & 86.6 \\
\bs PointGPT~\cite{PointGPT}$^\dagger$ & 360.5 & 67.7 & 93.4 & 94.7 & {98.0 {\small $\pm$ 1.9}} & 99.0 {\small $\pm$ 1.0} & 94.1 {\small $\pm$ 3.9} & 96.1 {\small $\pm$ 2.8} & 84.7 & 86.4 \\
\br PointMamba~\cite{PointMamba} & 12.3 & 3.6 & 89.31 & 93.6 & 95.0 {\small $\pm$ 2.3} & 97.3 {\small $\pm$ 1.8} & 91.4 {\small $\pm$ 3.3} & 92.8 {\small $\pm$ 4.0} & 84.4 & 86.0 \\
\br Mamba3D~\cite{Mamba3D} & 16.9 & 3.9 & {92.05} & {94.7} & {96.4 {\small $\pm$ 2.2}} & {98.2 {\small $\pm$ 1.2}} & {92.4 {\small $\pm$ 4.1}} & {95.2 {\small $\pm$ 2.9}} & 84.1 & 85.7 \\
\bs ReCon++~\cite{ReCon++}$^\dagger$ & 657.2 & - & {95.25} & {94.8} & 98.0 {\small $\pm$ 2.3} & {99.5 {\small $\pm$ 0.8}} & {94.5 {\small $\pm$ 4.1}} & {96.5 {\small $\pm$ 3.0}} & - & - \\
\midrule[0.4pt]
\multicolumn{11}{c}{\textit{with Pre-Training (Parameter-Efficient Fine-Tuning)}}\\
\midrule[0.4pt]
\bs ReCon~\cite{ReCon} & 22.1 & 4.8 & 90.63 & 94.1 & 97.3 {\small $\pm$ 1.9} & 98.9 {\small $\pm$ 1.2} & 93.3 {\small $\pm$ 3.9} & 95.8 {\small $\pm$ 3.0} & 84.52 & 86.1 \\
\quad + IDPT~\cite{IDPT} & 1.7 & 7.2 & 88.13 & 93.6 & 96.9 {\small $\pm$ 2.4} & 98.3 {\small $\pm$ 0.7} & {92.8 {\small $\pm$ 4.0}} & 95.5 {\small $\pm$ 3.2} & 83.66 & 85.7 \\
\quad + DAPT~\cite{DAPT} & 1.1 & 5.0 & 89.38 & 93.5 & 95.6 {\small $\pm$ 2.8} & 97.7 {\small $\pm$ 1.6} & 91.9 {\small $\pm$ 4.1} & 94.6 {\small $\pm$ 3.5} & 83.87 & 85.7 \\
\quad + PPT~\cite{PPT} & 1.1 & 11.3 & 89.52 & 93.8 & 97.0 {\small $\pm$ 2.7} & 98.7 {\small $\pm$ 1.6} & 92.2 {\small $\pm$ 5.0} & 95.6 {\small $\pm$ 2.9} & 84.23 & 85.6 \\
\quad + PointGST~\cite{PointGST} & 0.6 &  4.8 & 89.49 & 93.6 & {96.2 {\small $\pm$ 2.7}} & {98.2 {\small $\pm$ 1.1}} & {92.1 {\small $\pm$ 4.4}} & {95.6 {\small $\pm$ 2.8}} & 83.98 & 85.8 \\
\rowcolor{linecolor} \quad + \textbf{\ours} & 1.3 & 4.8 & \textbf{92.85} & \textbf{94.7} & \textbf{97.1 {\small $\pm$ 2.0}} & \textbf{98.9 {\small $\pm$ 0.9}} & \textbf{93.1 {\small $\pm$ 3.8}} & \textbf{95.6 {\small $\pm$ 2.8}} & \textbf{84.42} & \textbf{86.0} \\
\midrule[0.4pt]
\br Mamba3D~\cite{Mamba3D} & 16.9 & 3.9 & {92.05} & {94.7} & {96.4 {\small $\pm$ 2.2}} & {98.2 {\small $\pm$1.2}} & {92.4 {\small $\pm$ 4.1}} & {95.2 {\small $\pm$ 2.9}} & 84.13 & 85.7 \\
\quad + IDPT~\cite{IDPT} & 1.7 & 6.2 & 87.34 & 93.2 & 95.9 {\small $\pm$ 2.3} & 97.6 {\small $\pm$ 1.5} & 91.8 {\small $\pm$ 4.6} & 94.9 {\small $\pm$ 3.2} & 82.98 & 85.3 \\
\quad + DAPT~\cite{DAPT} & 1.1 & 4.0 & 88.55 & 92.9 & 96.0 {\small $\pm$ 2.0} & 97.7 {\small $\pm$ 1.7} & 92.1 {\small $\pm$ 4.5} & 94.8 {\small $\pm$ 3.7} & 83.49 & 85.5 \\
\quad + PPT~\cite{PPT} & 1.1 & 8.5 & 87.61 & 92.87 & 96.2 {\small $\pm$ 2.7} & 97.7 {\small $\pm$ 1.8} & 91.7 {\small $\pm$ 4.7} & 94.7 {\small $\pm$ 3.6} & 83.19 & 85.4 \\
\quad + PointGST~\cite{PointGST} & 0.6 &  3.9 & 89.97 & 93.7 & 96.5 {\small $\pm$ 2.1} & 97.5 {\small $\pm$ 1.6} & 92.4 {\small $\pm$ 4.7} & 94.9 {\small $\pm$ 3.1} & 83.39 & 85.5 \\
\rowcolor{linecolor} \quad + \textbf{\ours} & 1.2 & 3.9 & \textbf{92.85} & \textbf{94.7} & \textbf{96.8 {\small $\pm$ 2.4}} & \textbf{98.3 {\small $\pm$ 1.8}} & \textbf{93.0 {\small $\pm$ 4.5}} & \textbf{95.4 {\small $\pm$ 3.0}} & \textbf{83.74} & \textbf{85.7} \\
\bottomrule[0.7pt]
\end{tabular}
}
\end{center}
\vspace{-20pt}
\end{table*}

\subsection{Monarch Sparse Tuning}
\label{sec:most}
Monarch Sparse Tuning (\textbf{\ours}) replaces dense update weight matrices with Point Monarch, leveraging its sparsity for efficient reparameterization during training. 
\ours reduces parameters while capturing local features, making it highly \textbf{generalizable} and \textbf{flexible}, ideal for 3D PEFT. 
During inference, the reparameterized matrices merge seamlessly with pretrained weights, maintaining the original structure and achieving \textit{zero inference overhead}.

\vspace{5pt}
\noindent\textbf{Generalizable.}
For any pretrained dense weight matrix \( \mathbf{W}_0 \in \mathbb{R}^{d \times d} \) and a dense update matrix \( \Delta \mathbf{W} \in \mathbb{R}^{d \times d} \) fine-tuned for a downstream task, \ours is formulated as:
\begin{align}
h = \mathbf{W} x &= \mathbf{W}_0 x + \Delta \mathbf{W} x, \\
&= \mathbf{W}_0 x + \mathbf{\textcolor{mambacolor}{K} L R \textcolor{mambacolor}{K}} x.
\end{align}
\ours reparameterizes the dense update matrix \( \Delta \mathbf{W} \) (with \( d^2 \) parameters) into Point Monarch, reducing it to \( {2d^2}/{b} \) parameters, where typically \( b \approx \sqrt{d} \) controls sparsity. 
Similar to LoRA~\cite{LoRA}, we initialize $\mathbf{L} = 0$ and $\mathbf{R} = \mathcal{N}(0, \sigma^2)$, starting with \( \Delta \mathbf{W} = 0 \). 
This reparameterization generalizes well across models with dense layers, in contrast to current methods tailoring prompts or adapters for Transformers.

In practice, Point Monarch can handle non-square weights, specifically \( \Delta \mathbf{W} \in \mathbb{R}^{d_{\text{in}} \times d_{\text{out}}} \) where \( d_{\text{in}} \neq d_{\text{out}} \), allowing adaptation to various models and tasks. 
Assuming \( d_{\text{in}} < d_{\text{out}} \), we define \( \mathbf{R} \in \mathbb{R}^{d_{\text{in}} \times d_{\text{in}}} \) and \( \mathbf{L} \in \mathbb{R}^{d_{\text{in}} \times d_{\text{out}}} \). 
This implies that diagonal blocks can be rectangular matrices. 
Moreover, we can adjust the size of \( b \) based on model size or task complexity to control the parameter scaling.

\vspace{5pt}
\noindent\textbf{Flexible.}
\ours is also highly flexible. 
Replacing dense layers directly with Point Monarch significantly reduces parameters during both training and inference, sparsifying the entire model rather than just the update weights:
\begin{equation}
 \textit{Sparse Training:} \qquad  h = \mathbf{W} x = \mathbf{\textcolor{mambacolor}{K} L R \textcolor{mambacolor}{K}} x.
\end{equation}

Moreover, \ours is orthogonal to common matrix decompositions such as Low-rank and Kronecker~\cite{Kronecker}, allowing further parameter reduction when combined.
We explore some of these variants in \cref{sec:variants}.

\subsection{Multi-layer Feature Fusion}
\label{sec:fusion}
Our parameter-free fusion strategy aggregates features from multiple layers, avoiding bottlenecks between the pretrained backbone and the task header, boosting downstream tasks, defined as:
\begin{equation}
    \mathbf{x}_{\text{out}} = \sum\nolimits_{i=1}^{L} 2^{i-1} \cdot \text{MixPool}(\mathbf{x}_i).
\end{equation}
Here, \( \mathbf{x}_i \in \mathbb{R}^{G \times C} \) (\( i=1, \ldots, L \)) represents one token sequence from \( L \) selected layers. For a 12-layer Transformer or Mamba model, we select \( L=3 \) layers: the 4th, 8th, and 12th. 
Each sequence undergoes a mixture of MaxPooling and MeanPooling over its sequence length \( G \), followed by a weighted sum $2^{i-1}$ (\( i=1, \ldots, L \)), yielding \( \mathbf{x}_{\text{out}} \in \mathbb{R}^{1 \times C} \). 
Finally, we concatenate $\mathbf{x}_{\text{out}}$ with the class token from the last layer, which serves as input to the task header.

\begin{table}[t!]
\renewcommand{\arraystretch}{1.02}
\centering
\caption{
Comparison results of large-scale indoor scene semantic segmentation on S3DIS Area 5~\cite{S3DIS16}.
}
\vspace{-5pt}
\label{tab:s3dis}
\Large
\resizebox{1.0\linewidth}{!}{
\begin{tabular}{lcccc}
\toprule[0.7pt]
Method & Params. (M) & input & mAcc (\%) $\uparrow$ & mIoU (\%) $\uparrow$ \\
\midrule[0.4pt]
\multicolumn{5}{c}{\textit{Supervised Learning Only}} \\
\midrule[0.4pt]
\bh PointNet$^\dagger$~\citep{PointNet} & 3.6 & xyz+rgb & 49.0 & 41.1 \\
\bh PointCNN$^\dagger$~\citep{PointNet} & 3.6 & xyz+rgb & 63.9 & 57.3 \\
\bh PointNet++$^\dagger$~\citep{PointNet++} & {1.0} & xyz+rgb & 67.1 & 53.5 \\
\bh PCT$^\dagger$~\cite{PCT} & 2.9 & xyz+rgb & 67.7 & {61.3} \\
\bs Transformer~\cite{AttentionIsAllYouNeed} & 27.0 & xyz & {68.6} & 60.0  \\
\br Mamba3D~\cite{Mamba3D} & {21.9} & xyz & 67.8 & 58.0  \\
\midrule[0.4pt]
\multicolumn{5}{c}{\textit{Supervised Learning Only ({Sparse Training})}}\\
\midrule[0.4pt]
\rowcolor{linecolor1} \bs \textbf{\ours} {\large (Transformer)} & 5.3 & xyz & \textbf{64.7} & \textbf{51.4}  \\
\rowcolor{linecolor} \br \textbf{\ours} {\large (Mamba3D)} & 5.9 & xyz & \textbf{67.6} & \textbf{57.1}  \\
\midrule[0.4pt]
\multicolumn{5}{c}{\textit{with Pretraining (Full Finetuning)}} \\
\midrule[0.4pt]
\bs Point-BERT~\cite{PointBERT} & 27.0 & xyz & 69.7 & 60.5  \\
\bs Point-MAE~\cite{PointMAE} & 27.0 & xyz & {69.9} & {60.8}  \\
\bs ReCon~\cite{ReCon} & 27.0 & xyz & 69.7 & {60.8}  \\
\br Mamba3D~\cite{Mamba3D} & {21.9} & xyz & 68.7 & {60.7}  \\
\midrule[0.4pt]
\multicolumn{5}{c}{\textit{with Pretraining (Efficient Finetuning)}} \\
\midrule[0.4pt]
\bs ReCon~\cite{ReCon} & 27.0 & xyz & 69.7 & 60.8  \\
\quad + Linear Probing & 5.2 & xyz &  64.3 & 51.2  \\
\quad + IDPT~\cite{IDPT} & 5.6 & xyz &  62.9 & 50.5   \\
\quad + DAPT~\cite{DAPT} & 5.6 & xyz & 66.3 & 56.3  \\
\quad + PPT~\cite{PPT} & 5.6 & xyz & 65.6 & 54.8  \\ 
\quad + PointGST~\cite{PointGST} & 5.6 & xyz & 67.8 & 57.9  \\
\rowcolor{linecolor} \quad + \textbf{\ours} & 5.8 & xyz & \textbf{68.8} & \textbf{58.9}  \\
\bottomrule[0.7pt]
\end{tabular}
}
\vspace{-10pt}
\end{table}

\section{Experiments}

\subsection{3D Representation Learning Backbones}
We assess the generalization and effectiveness of \ours using five distinct point cloud pretraining model backbones:

\textbf{Point-MAE}~\cite{PointMAE}: A Transformer (22.1M) pretrained with masked point modeling (MPM) on ShapeNet~\cite{ShapeNet15} (\(\sim\)50K). 
 
\textbf{I2P-MAE}~\cite{I2P-MAE}: A U-Net-like hierarchical architecture (15.3M) with multimodal pretraining.
 
\textbf{ReCon}~\cite{ReCon}: A Transformer (22.1M) with multimodal pretraining on ShapeNet~\cite{ShapeNet15}.
 
\textbf{Mamba3D}~\cite{Mamba3D}: A Mamba model (16.9M) pretrained with MPM on ShapeNet~\cite{ShapeNet15}.
 
\textbf{PointGPT}~\cite{PointGST}: A large Transformer (360.5M) pretrained with MPM on a large mixed dataset (\(\sim\)300K).

\subsection{Implementation Details}
We follow the full fine-tuning experimental setup but freeze the pretrained backbone, making only Point Monarch and the task header trainable. 
Unless otherwise specified, we use Point Monarch with a block count \( b = 16 \) and \( K \)-Rectify with \( K = 4 \).
To optimize spectral norm at initialization and stabilize training while scaling \( b \), we initialize \( \mathbf{L} = 0 \) and \( \mathbf{R} = \mathcal{N}(0, \sigma^2) \), where \( \sigma = \sqrt{\min(d_{\text{in}}, d_{\text{out}})/(d_{\text{in}})^2} \), based on scaling laws for structured matrices~\cite{Compute_Better_Spent}. 
To reduce hyperparameter tuning, we set \( \lambda \) in \( K \)-Rectify as learnable. 
We reparameterize the linear projections in Transformer’s Attention~\cite{AttentionIsAllYouNeed}, Mamba’s SSM~\cite{Mamba}, and the FFN~\cite{AttentionIsAllYouNeed} across all models.
All experiments are conducted on a single RTX 4090 GPU. 
See Appendix for more details.

\subsection{Comparison on Downstream Tasks}
\textbf{3D Real-World / Synthetic Object Recognition.}
We evaluate shape classification on the challenging PB\_50\_RS variant of the real-world ScanObjectNN~\cite{ScanObjectNN19} and the synthetic ModelNet40~\cite{ModelNet15} datasets. 
As shown in ~\cref{tab:peft_main}, we compare two types of PEFT methods: additive (e.g., VPT~\cite{VPT22}, IDPT~\cite{IDPT}, DAPT~\cite{DAPT}) and selective/reparameterization (e.g., BitFit~\cite{BitFit}, LoRA~\cite{LoRA}, and our \ours). 
\ours consistently delivers the best performance across all models. 
With only 0.8M (3.6\%) trainable parameters (\( b=32 \)), even Point-MAE~\cite{PointMAE} can surpass full fine-tuning by 6.77\% on ScanObjectNN and 0.24\% on ModelNet40. 
Furthermore, with \( b=8 \), \ours achieves an average improvement of 3.96\% and 1.52\% across five models compared to full fine-tuning, while other PEFT methods like PPT~\cite{PPT} and LoRA~\cite{LoRA} fall short. 
In particular, \ours allows PointGPT~\cite{PointGPT} to reach 97.50\% accuracy on ScanObjectNN and 96.23\% on ModelNet40, setting new state-of-the-art results. 
These results underscore the effectiveness and strong generalizability of \ours.

\vspace{3pt}
\noindent\textbf{Few-Shot Learning.}
We follow previous work~\cite{PointSSL20} to conduct few-shot learning experiments on ModelNet40~\cite{ModelNet15}, using an $n$-way, $m$-shot setting with $n \in \{5, 10\}$ and $m \in \{10, 20\}$. 
As shown in ~\cref{tab:cls_main}, \ours fine-tuned on Mamba3D~\cite{Mamba3D} and ReCon~\cite{ReCon} outperforms existing 3D PEFT methods, demonstrating its effectiveness. 
Furthermore, when using sparse training---replacing dense layers with Point Monarch and training from scratch---\ours boosts the performance of both Transformer~\cite{AttentionIsAllYouNeed} and Mamba3D. 
For Transformer, \ours achieves an average improvement of 1.35\% acc. while reducing the parameter count to 1/5.

\vspace{3pt}
\noindent\textbf{3D Part Segmentation.}
We conduct part segmentation experiments on ShapeNetPart~\cite{ShapeNetPart16}, reporting both average instance IoU (mIoU$_I$) and average category IoU (mIoU$_C$), as shown in ~\cref{tab:cls_main}. 
\ours fine-tuned on Mamba3D~\cite{Mamba3D} and ReCon~\cite{ReCon} surpasses existing 3D PEFT methods and achieves results comparable to full fine-tuning. 
Even with sparse training, Mamba3D and Transformer~\cite{AttentionIsAllYouNeed} achieve mIoU$_C$ of 82.93\% and 83.11\%, respectively. 
This fine-grained segmentation experiment further highlights \ours’s ability to capture fine-grained local features, enhancing its capability to process point clouds.

\begin{figure}[t]
    \begin{center}
    \includegraphics[width=1.0\linewidth]{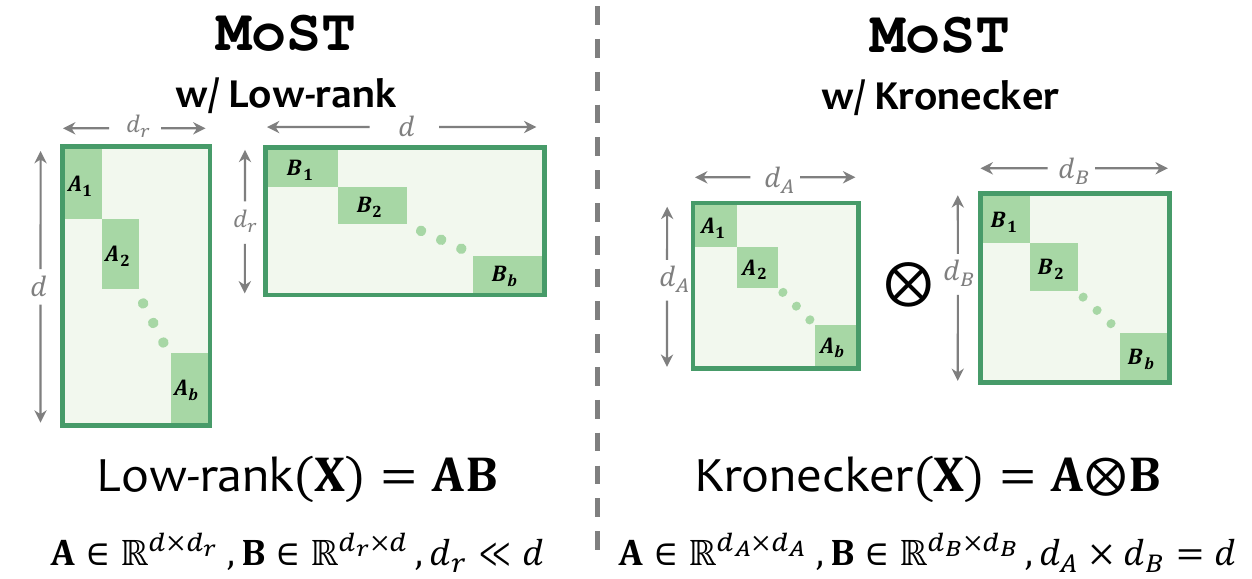}
    \vspace{-15pt}
    \caption{Variants with Low-rank or Kronecker decomposition. 
    }\label{fig:variant}
    \end{center}
    \vspace{-15pt}
\end{figure}

\vspace{3pt}
\noindent\textbf{3D Scene Semantic Segmentation. }
To further verify \ours's effectiveness on large-scale scene datasets, we conduct semantic segmentation experiments on the S3DIS Area 5~\cite{S3DIS16}, with results shown in ~\cref{tab:s3dis}. 
In this scene segmentation task, most fine-tuning parameters are concentrated in the segment header, resulting in similar parameter counts across methods. 
Nevertheless, \ours achieves 68.8\% mAcc and 58.9\% mIoU, outperforming other 3D PEFT methods and coming closest to full fine-tuning performance. 
When using \ours for sparse training, the Transformer~\cite{AttentionIsAllYouNeed} requires only 1/4 of the parameters to achieve 64.7\% mAcc and 51.4\% mIoU, while Mamba3D\cite{Mamba3D} achieves 67.6\% mAcc and 57.1\% mIoU. 
This experiment demonstrates \ours’s generalization and effectiveness on large-scale scenes.

\begin{table}[t]
\renewcommand{\arraystretch}{1.05}
\centering
\caption{Results of \ours variants in different combinations from \cref{fig:variant} with PointGPT~\cite{PointGPT}. $^\dagger$: joint Low-rank decomposition.}
\vspace{-5pt}
\label{tab:variant}
\resizebox{1.0\linewidth}{!}{
\begin{tabular}{c|cccc}
\toprule[0.7pt]
Method & Matrix $\mathbf{L}$  & Matrix $\mathbf{R}$ & Params. (M) & PB\_50\_RS (\%) \\
\midrule[0.4pt]
I & Low-rank & Low-rank & 2.1 \dperc{0.6} & 97.02 \dminus{-0.17} \\
II & Kronecker & Kronecker & 0.8 \dperc{0.2} & 96.08 \dminus{-1.11} \\
III & Low-rank$^\dagger$ & Low-rank$^\dagger$ & 1.6 \dperc{0.4} & 96.88 \dminus{-0.31} \\
\midrule[0.4pt]
IV & $\mathbf{L}$ & Low-rank & 2.3 \dperc{0.6} & 97.09 \dminus{-0.10} \\
V & $\mathbf{L}$ & Kronecker & 1.7 \dperc{0.5} & 97.09 \dminus{-0.10} \\
VI & Low-rank & Kronecker & 1.4 \dperc{0.4} & 96.18 \dminus{-1.01} \\
\midrule[0.4pt]
\rowcolor{linecolor} \textbf{\ours}$_{b=32}$ & $\mathbf{L}$ & $\mathbf{R}$ & 2.5 \dperc{0.7} & \textbf{97.19} \dplus{+0.00} \\
\bottomrule[0.7pt]
\end{tabular}
}
\vspace{-10pt}
\end{table}
\begin{figure*}[t]
    \begin{center}
    \vspace{-4pt}
    \includegraphics[width=1.0\linewidth]{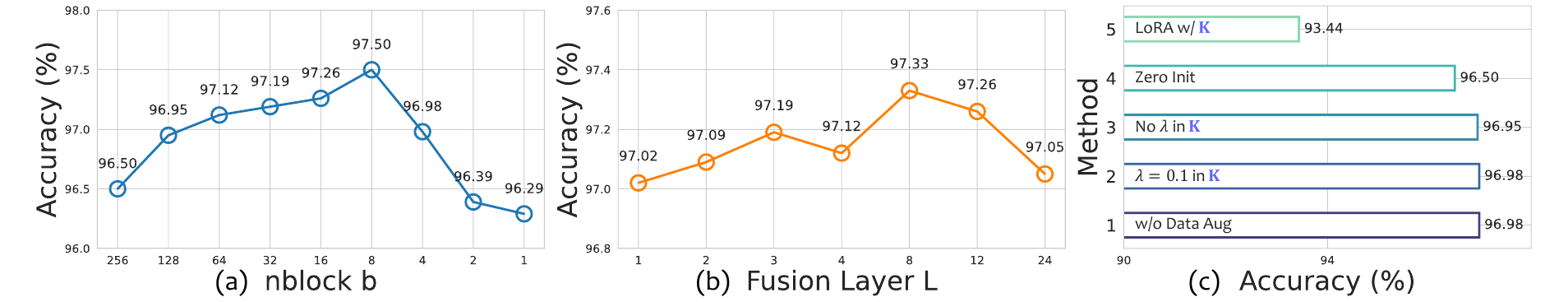}
    \vspace{-15pt}
    \caption{
    \textbf{Ablation of parameters \( b \), \( L \), and others}. We report classification results (\%) on ScanObjectNN PB\_50\_RS using PointGPT~\cite{PointGPT}.
    }\label{fig:other_ablation}
    \vspace{-20pt}
    \end{center}
\end{figure*}

\subsection{Variants}
\label{sec:variants}
We explore \ours variants that combine Low-rank and Kronecker~\cite{Kronecker} decompositions, as shown in \cref{fig:variant}. 
Both block-diagonal matrix \( L \) and \( R \) can be decomposed to two sub-matrices, or we can treat \( L \) and \( R \) as the results of a joint Low-rank decomposition.
Including hybrid methods, we verify 6 variants, with results in \cref{tab:variant}. 
The Low-rank and Kronecker decompositions reduce parameters to 2.1M (0.6\%) and 0.8M (0.2\%) with only a slight performance drop. 
Combined, they minimize parameters to 1.4M (0.4\%) and achieve an accuracy of 96.18\%, highlighting the flexibility of \ours in balancing performance and efficiency.

\begin{table}[t!]
\renewcommand{\arraystretch}{1.05}
\centering
\caption{Comparisons of PEFT vs. 3D PEFT on the PB\_50\_RS~\cite{ScanObjectNN19} using Point-MAE~\cite{PointMAE}. \texttt{NO}: no inference overhead.
}
\vspace{-5pt}
\label{tab:1d_2d_3d_peft}
\large
\resizebox{1.0\linewidth}{!}{
\begin{tabular}{lcccc}
\toprule[0.7pt]
 Method  & Domain & \texttt{NO} & \#P (M) & SONN (\%) \\
\midrule[0.4pt]
Point-MAE~\cite{PointMAE}   & - & - & 22.1 \dperc{100} & 85.18 \dplus{+0.00} \\
\quad  + Linear probing &- & - & 0.3 \dperc{1.4} & 75.99  \dminus{-9.19}\\
 \midrule[0.4pt]
\quad  + Adapter~\cite{Adapter}  &1D & \ding{55} & 0.9 \dperc{4.1} & 83.93 \dminus{-1.25} \\
\quad  + Perfix tuning~\cite{Prefix-Tuning}  &1D & \ding{55} &0.7 \dperc{3.2} & 77.72 \dminus{-7.46} \\
\quad  + BitFit~\cite{BitFit}  &1D & \textbf{\textcolor{newgreen}{\ding{51}}} &0.3 \dperc{1.4} &  82.62 \dminus{-2.56}    \\
\quad  + LoRA~\cite{LoRA}  &1D & \textbf{\textcolor{newgreen}{\ding{51}}} & 0.9 \dperc{4.1}&    82.76 \dminus{-2.42}   \\
\quad  + DePT~\cite{DePT} &1D & \ding{55}  &0.3 \dperc{1.4} &    79.70 \dminus{-5.48}\\
\quad  + FourierFT~\cite{FourierFT} &1D & \textbf{\textcolor{newgreen}{\ding{51}}}  &0.3 \dperc{1.4} &    78.57 \dminus{-6.61}\\
  \midrule[0.4pt]
\quad  + VPT-Deep~\cite{VPT22}&2D & \ding{55}  &0.4 \dperc{1.8} &    81.09 \dminus{-4.09}\\
\quad  + AdaptFormer~\cite{AdaptFormer} &2D & \ding{55}  &0.9 \dperc{4.1}  &   83.45 \dminus{-1.73} \\
 \quad + SSF~\cite{SSF} &2D & \ding{55}   &0.4  \dperc{1.8} &   82.58 \dminus{-2.60} \\
\quad  + FacT~\cite{FacT} &2D & \textbf{\textcolor{newgreen}{\ding{51}}}  & 0.5 \dperc{2.3} &    78.76 \dminus{-6.42} \\
\quad  + BI-AdaptFormer~\cite{BI-AdaptFormer} &2D & \ding{55}   &0.4 \dperc{2.3} &   83.66 \dminus{-1.52} \\
\quad  + SCT~\cite{SCT} &2D & \ding{55}   &0.3 \dperc{1.4} &     80.40 \dminus{-4.78} \\
  \midrule[0.4pt]
\quad  + IDPT~\cite{IDPT} &3D & \ding{55}  & 1.7 \dperc{7.7} &  88.34 \dplus{+3.16} \\
\quad  + DAPT~\cite{DAPT} &3D  & \ding{55} & 1.1 \dperc{5.0} &   88.27 \dplus{-1.25} \\
\quad  + PPT~\cite{PPT} &3D & \ding{55}  & 1.1 \dperc{5.0} &   89.00 \dplus{+3.09} \\
\quad  + PointGST~\cite{PointGST} &3D & \ding{55}& 0.6 \dperc{2.7} &   89.3  \dplus{+4.12} \\
\rowcolor{linecolor} \quad  + \textbf{\ours}$_{b=32}$ &3D & \textbf{\textcolor{newgreen}{\ding{51}}}& 0.8 \dperc{3.6} &   \textbf{91.95} \dplus{+6.77} \\
\bottomrule[0.7pt]
\end{tabular}
}
\vspace{-10pt}
\end{table}

\subsection{PEFT vs. 3D PEFT}
\cref{tab:1d_2d_3d_peft} compares various PEFT methods using Point-MAE~\cite{PointMAE} on the ScanObjectNN PB\_50\_RS dataset~\cite{ScanObjectNN19}, including those for text, images, and 3D point clouds. 
Due to the inherent irregularity of point clouds, traditional PEFT methods for text and images perform poorly, with the best Adapter~\cite{Adapter} achieving only 83.93\% accuracy. 
In contrast, 3D PEFT methods reach at least 88.27\% accuracy, with our \ours achieving 91.95\% using only 0.9M fine-tuned parameters, significantly surpassing full fine-tuning.

\subsection{Ablation Study}

\textbf{Components.} 
We perform ablation studies on each component using PointGPT~\cite{PointGPT} on the ScanObjectNN PR\_50\_RS classification dataset~\cite{ScanObjectNN19}, as shown in \cref{tab:ablation}. 
Removing \( \textcolor{mambacolor}{\mathbf{K}} \) results in a 2.88\% accuracy drop, while eliminating Point Monarch entirely causes a 6.8\% decrease.
The Feature Fusion shows that both $2^{i-1}$ weighting and MixPool enhance accuracy. 
Using only Monarch yields 92.68\% acc., significantly lower than Point Monarch's 96.81\%. 
Notably, accuracy drops to 90.90\% when removing Monarch, which requires only 0.7M parameters to finetune (0.2\%). 
These results validate the effectiveness of \ours and underscore Point Monarch's key role in capturing local information.

\vspace{3pt}
\noindent\textbf{Parameters.} 
We evaluate the impact of block number \( b \) in Point Monarch and layer number \( L \) in Feature Fusion on model performance, as shown in \cref{fig:other_ablation}(a-b). 
Results indicate that accuracy initially increases with \( b \) and peaks at \( b = 8 \) before declining. 
At \( b = 256 \), the model still achieves 96.5\% acc. with only 0.9M parameters, demonstrating Point Monarch's expressiveness. 
For \( L \) in Feature Fusion, \( L = 3 \) performs well, while peaking at \( L = 8 \) with 97.33\% acc., confirming the effectiveness of multi-layer feature fusion.

\vspace{3pt}
\noindent\textbf{Others.} 
Ablations on \( \lambda \), initialization, LoRA with \( \textcolor{mambacolor}{\mathbf{K}} \), and data augmentation~\cite{ACT23} are shown in \cref{fig:other_ablation}(c). 
A learnable \( \lambda \) yields optimal results, while 0-initializing both \( \mathbf{L} \) and \( \mathbf{R} \) decreases accuracy. 
Incorporating \( \textcolor{mambacolor}{\mathbf{K}} \) enhances LoRA, though it remains less expressive than Point Monarch. Rotation as data augmentation~\cite{ACT23} provides limited improvements.

\begin{table}[t!]
\renewcommand{\arraystretch}{1.05}
\centering
\caption{Ablation study using PointGPT~\cite{PointGPT} on ScanObjectNN~\cite{ScanObjectNN19}, where ``w/o Monarch" denotes fine-tuning only the task header.}
\vspace{-5pt}
\label{tab:ablation}
\resizebox{1.0\linewidth}{!}{
\begin{tabular}{ccccccc}
\toprule[0.7pt]
\multicolumn{3}{c}{Point Monarch} & \multicolumn{2}{c}{Feature Fusion} & \multirow{2}{*}[-0.5ex]{\#P} & \multirow{2}{*}[-0.5ex]{PB\_50\_RS (\%)} \\
\cmidrule(lr){1-3} \cmidrule(lr){4-5} 
\( \textcolor{mambacolor}{\mathbf{K}} \) & IDW & Monarch & $2^{i-1}$ & MixPool & &  \\
\midrule[0.4pt]
 \checkmark & - & \checkmark & \checkmark & \checkmark & 2.5 \dperc{0.7} & 96.77 \dminus{-0.42} \\
\checkmark & \checkmark & - & \checkmark & \checkmark & 0.7 \dperc{0.2} & 90.90 \dminus{-6.29} \\
 - & - & \checkmark & \checkmark & \checkmark & 2.5 \dperc{0.7} & 94.31 \dminus{-2.88} \\
 - & - & - & \checkmark & \checkmark & 0.7 \dperc{0.2} & 90.39 \dminus{-6.80} \\ 
\midrule[0.4pt]
 \checkmark & \checkmark & \checkmark & - & \checkmark & 2.5 \dperc{0.7} & 97.05 \dminus{-0.14} \\
 \checkmark & \checkmark & \checkmark & \checkmark & - & 2.5 \dperc{0.7} & 96.91 \dminus{-0.28} \\
 \checkmark & \checkmark & \checkmark & - & - & 2.5 \dperc{0.7} & 96.81 \dminus{-0.38} \\
\midrule[0.4pt]
 - & - & \checkmark & - & \checkmark & 2.5 \dperc{0.7} & 94.03 \dminus{-3.16} \\
 - & - & \checkmark & \checkmark & - & 2.5 \dperc{0.7} & 93.93 \dminus{-3.26} \\
 - & - & \checkmark & - & - & 2.5 \dperc{0.7} & 92.68 \dminus{-4.51} \\
\midrule[0.4pt]
\rowcolor{linecolor}  \checkmark & \checkmark & \checkmark & \checkmark & \checkmark & 2.5 \dperc{0.7} & \textbf{97.19} \dplus{+0.00} \\
\bottomrule[0.7pt]
\end{tabular}
}
\vspace{-10pt}
\end{table}

\section{Conclusion}

This paper explores reparameterization-based PEFT in 3D representation learning. 
We introduce \ours, the first reparameterization-based 3D PEFT method tailored for point clouds. 
At its core, we propose Point Monarch, a novel structured matrix family that uses simple linear transformations to capture local geometric features. 
\ours reparameterizes dense update weight matrices with local-aware, sparse Point Monarch matrices, achieving state-of-the-art results while maintaining generalizability. 
Additionally, \ours can combine with matrix decompositions like Low-rank and Kronecker for further parameter reduction. 
We hope this work will inspire advances in fine-tuning large-scale 3D models.

\section*{Acknowledgements.}
This work was supported by the China National Natural Science Foundation No. 62202182.
%



{
    \small
    \bibliographystyle{ieeenat_fullname}
    \bibliography{main}
}

\end{document}